\documentclass{article}

\usepackage[preprint]{neurips_2021}

\usepackage[utf8]{inputenc} 
\usepackage[T1]{fontenc}    
\usepackage[breaklinks=true]{hyperref}       
\usepackage{url}            
\usepackage{booktabs}       
\usepackage{amsfonts}       
\usepackage{nicefrac}       
\usepackage{microtype}      
\usepackage[dvipsnames]{xcolor}         
\usepackage{algorithm}
\usepackage[noend]{algpseudocode}
\usepackage{algorithmicx}
\usepackage{amsmath,amssymb,amsfonts,amsthm}
\newcommand{\norm}[1]{\left\lVert#1\right\rVert}

\newcommand{\bdelta}{\boldsymbol{\delta}}
\newcommand{\Zero}{\boldsymbol{0}}
\newcommand{\I}{\boldsymbol{\mathrm{I}}}
\newcommand{\x}{\boldsymbol{\mathrm{x}}}
\newcommand{\A}{\boldsymbol{\mathrm{A}}}
\newcommand{\B}{\boldsymbol{\mathrm{B}}}
\newcommand{\y}{\boldsymbol{\mathrm{y}}}
\newcommand{\z}{\boldsymbol{\mathrm{z}}}
\newcommand{\w}{\boldsymbol{\mathrm{w}}}
\newcommand{\wb}{\boldsymbol{\Bar{\mathrm{w}}}}
\newcommand{\dr}{\mathrm{d}}

\newcommand{\thicktilde}[1]{\mathbf{\Tilde{\text{$#1$}}}}

\newcommand{\xp}{\boldsymbol{\mathrm{x}}'}
\newcommand{\xheun}{\thicktilde{\boldsymbol{\mathrm{x}}}}
\newcommand{\xpp}{\boldsymbol{\mathrm{x}}''}

\usepackage{graphicx}
\usepackage{subcaption}
\usepackage{caption}

\algblock{ParFor}{EndParFor}
\algnewcommand\algorithmicparfor{\textbf{parfor}}
\algnewcommand\algorithmicpardo{\textbf{do}}
\algnewcommand\algorithmicendparfor{\textbf{end\ parfor}}
\algrenewtext{ParFor}[1]{\algorithmicparfor\ #1\ \algorithmicpardo}
\algrenewtext{EndParFor}{\algorithmicendparfor}

\newcommand\blfootnote[1]{%
  \begingroup
  \renewcommand\thefootnote{}\footnote{#1}%
  \addtocounter{footnote}{-1}%
  \endgroup
}

\title{Gotta Go Fast When Generating Data with Score-Based Models}

 \author{
 Alexia Jolicoeur-Martineau\\
 Department of Computer Science\\
 University of Montreal
 \And 
 Ke Li\textsuperscript{$\dagger$} \\
 Department of Computer Science\\
 Simon Fraser University
 \And 
 Rémi Piché-Taillefer\textsuperscript{$\dagger$} \\
 Department of Computer Science\\
 University of Montreal
 \And 
 Tal Kachman\textsuperscript{$\dagger$} \\
 Department of Artificial Intelligence \\
 Radboud University \& Donders Institute
\And
 Ioannis Mitliagkas \\
 Department of Computer Science\\
 University of Montreal
}
\begin{document}
\blfootnote{$\dagger$ Equal contribution}

\maketitle

\begin{abstract}

   Score-based (denoising diffusion) generative models have recently gained a lot of success in generating realistic and diverse data. These approaches define a forward diffusion process for transforming data to noise and generate data by reversing it (thereby going from noise to data). Unfortunately, current score-based models generate data very slowly due to the sheer number of score network evaluations required by numerical SDE solvers. 
   
   In this work, we aim to accelerate this process by devising a more efficient SDE solver. Existing approaches rely on the Euler-Maruyama (EM) solver, which uses a fixed step size. We found that naively replacing it with other SDE solvers fares poorly - they either result in low-quality samples or become slower than EM. To get around this issue, we carefully devise an SDE solver with adaptive step sizes tailored to score-based generative models piece by piece. Our solver requires only two score function evaluations, rarely rejects samples, and leads to high-quality samples. Our approach generates data 2 to 10 times faster than EM while achieving better or equal sample quality. For high-resolution images, our method leads to significantly higher quality samples than all other methods tested. Our SDE solver has the benefit of requiring no step size tuning.
   
   Code is available on https://github.com/AlexiaJM/score\_sde\_fast\_sampling.
 
\end{abstract}

\section{Introduction}

Score-based generative models \citep{song2019generative, song2020improved, ho2020denoising, jolicoeur2020adversarial, song2020score, hmc2021remipt} have been very successful at generating data from various modalities, such as images \citep{ho2020denoising, song2020score}, audio \citep{chen2020wavegrad,kong2020diffwave, mittal2021symbolic, kameoka2020voicegrad}, and graphs \citep{niu2020permutation}. They have further been used effectively for super-resolution \citep{saharia2021image,kadkhodaie2020solving}, inpainting \citep{kadkhodaie2020solving, song2020improved}, source separation \citep{jayaram2020source}, and image-to-image translation \citep{sasaki2021unit}. In most of these applications, score-based models achieved superior performances in terms of quality and diversity than the historically dominant Generative Adversarial Networks (GANs) \citep{GAN}.

\begin{figure}[H] 
    \centering
    \includegraphics[width=1\linewidth]{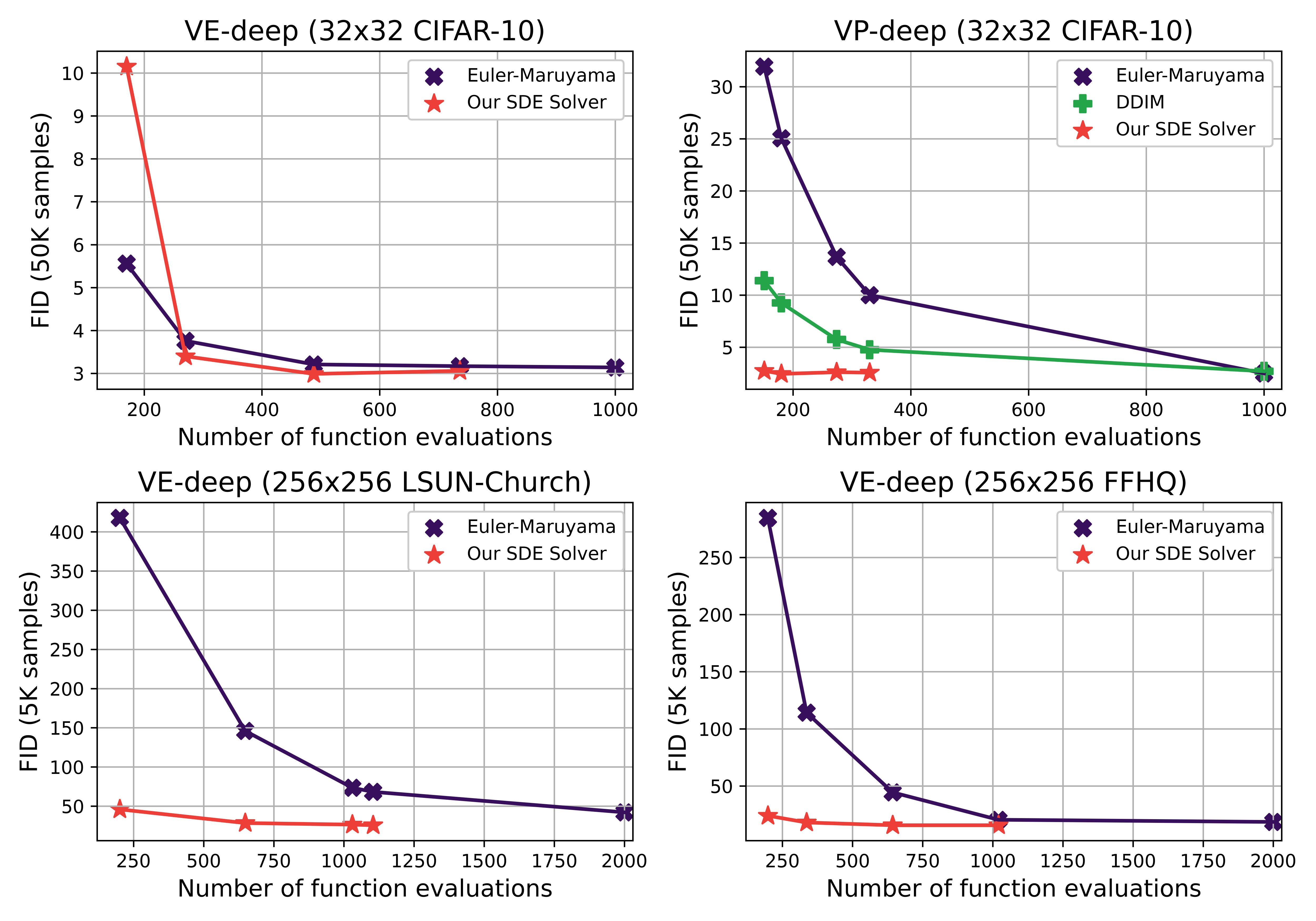}
    \caption{Comparison between our novel SDE solver at various 
    values of error tolerance and Euler-Maruyama for an equal computational budget. We measure speed through the Number of Function Evaluations (NFE) and the quality of the generated images through the Fréchet Inception Distance (FID; lower is better). See Table \ref{tab:table1}-\ref{tab:table2} for more details.} 
  \label{fig:flagship} 
\end{figure}


In Score-based approaches, a diffusion process progressively transforms real data into Gaussian noise. Then, the process is reversed in order to generate real data from Gaussian noise. Reversing the process requires the score function, which is estimated with a neural network (known as a score network). Although very powerful, score-based models generate data through an undesirably long iterative process; meanwhile, other state-of-the-art methods such as GANs generate data from a single forward pass of a neural network. Increasing the speed of the generative process is thus an active area of research.

\citet{chen2020wavegrad} and \citet{san2021noise} proposed faster step size schedules for VP diffusions that still yield relatively good quality/diversity metrics. Although fast, these schedules are arbitrary, require careful tuning, and the optimal schedules will vary from one model to another.

\citet{block2020fast} proposed generating data progressively from low to high-resolution images and show that the scheme improves speed. Similarly, \citet{nichol2021improved} proposed generating low-resolution images and then upscale them since generating low-resolution images is quicker. 
They further suggested to accelerate VP-based models by learning dimension-specific noise rather than assuming equal noise everywhere.
Note that these methods do not affect the data generation algorithm and would thus be complementary to our methods.

\citet{song2020score} and \citet{song2020denoising} proposed removing the noise from the data generation algorithm and solve an Ordinary Differential Equation (ODE) rather than a Stochastic Differential Equation (SDE); they report being able to converge much faster when there is no noise. Although it improves the generation speed, \citet{song2020score} report obtaining lower-quality images when using the ODE formulation for the VE process \citep{song2020score}. 
We will later show that our SDE solver generally leads to better results than ODE solvers at similar speeds.

Thus, existing methods for acceleration often require considerable step size/schedule tuning (this is also true for the baseline approach) and do not always work for both VE and VP processes. To improve speed and remove the need for step size/schedule tuning, we propose to solve the reverse diffusion process using SDE solvers with adaptive step sizes. 

It turns out that off-the-shelf SDE solvers are ill-suited for generative modeling and exhibit (1) divergence, (2) slower data generation than the baseline, or (3) significantly worse quality than the baseline (see Appendix \ref{sec:old}). 
This can be attributed to distinct features of the SDEs that arise in score-based generative models that set them apart from the SDEs traditionally considered in the numerical SDE solver literature, namely: (1) the codomain of the unknown function is extremely high-dimensional, especially in the case of image generation;
(2) evaluating the score function is computationally expensive, requiring a forward pass of a large mini-batch through a large neural network; (3) the required precision of the solution is smaller than usual because we are satisfied as long as the error is not perceptible (e.g., one RGB increment on an image).

We devise our own SDE solver with these features in mind, resulting in an algorithm that can get around the problems encountered by off-the-shelf solvers. To address high dimensionality, we use the $\ell_2$ norm rather than the $\ell_\infty$ norm to measure the error across different dimensions to prevent a single pixel from slowing down the solver. To address the cost of score function evaluations while still obtaining high precision, we (1) take the minimum number of score function evaluations needed for adaptive step sizes (two evaluations), and (2) use extrapolation to get high precision at no extra cost.
To take advantage of the reduced requirement for precision, we set the absolute tolerance for the error according to the range of RGB values. 
 

Our main contribution is a new SDE solver tailored to score-based generative models with the following benefits:
\begin{itemize}
 \item Our solver is much faster than the baseline methods, i.e. reverse-diffusion method with Langevin dynamics and Euler-Maruyama (EM);
 \item It yields higher quality/diversity samples than EM when using the same computing budget;
 \item It does not require any step size or schedule tuning;
 \item It can be used to quickly solve any type of diffusion process (e.g., VE, VP)
\end{itemize}

\section{Background}

\subsection{Score-based modeling with SDEs}

Let $\x(0) \in \mathbb{R}^d$ be a sample from the data distribution $p_{\text{data}}$. 
The sample is gradually corrupted over time through a Forward Diffusion Process (FDP), a common type of Stochastic Differential Equation (SDE): \begin{equation}\label{eq:forward}
\dr\x = f(\x,t)\dr t + g(t)\dr \w,
\end{equation}
where $f(\x,t): \mathbb{R}^d \times \mathbb{R} \to \mathbb{R}^d$ is the drift, $g(t): \mathbb{R} \to \mathbb{R}$ is the diffusion and $\w(t)$ is the Wiener process indexed by $t \in [0,1]$. Data points and their probability distribution evolve along the trajectories $\{ \x(t) \}_{t=0}^1$ and $\{p_t(\x) \}_{t=0}^1$ respectively, with $p_0 \equiv p_{\text{data}}$. 
The functions $f$ and $g$ are chosen such that $\x(1)$ be approximately Gaussian and independent from $\x(0)$. 
Inference is achieved by reversing this diffusion, drawing $\x(1)$ from its Gaussian distribution and solving the Reverse Diffusion Process (RDP) equal to:
\begin{equation}\label{eq:backward}
    \dr\x = \left[ f(\x,t)-g(t)^2\nabla_{\x} \log p_t(\x) \right] \dr t + g(t)\dr \wb,
\end{equation}
where $\nabla_{\x} \log p_t(\x)$ is 
referred to as the score
of the distribution at time $t$~\citep{hyvarinen2005estimation} and $\wb(t)$ is the Wiener process in which time flows backward~\citep{anderson1982reverse}.

One can observe from Equation \ref{eq:backward} that the RDP requires knowledge of the score (or $p_t$), which we do not have access to. Fortunately, it can be estimated by a neural network (referred to as the score network) by optimizing the following objective:
\begin{equation}\label{eqn:dsm}
\mathcal{L}(\theta) = \mathbb{E}_{ \x(t)\sim p(\x(t)|\x(0)), \x(0) \sim p_{\text{data}}}\left[ \frac{\lambda(t)}{2}\norm{s_{\theta}(\x(t),t) - \nabla_{\x(t)} \log p_t(\x(t)| \x(0)) }_2^2 \right],
\end{equation}
where $\lambda(t): \mathbb{R} \to \mathbb{R}$ is a weighting function generally chosen to be inversely proportional to: $$\mathbb{E} \left[ \norm{\nabla_{\x(t)} \log p_t(\x(t)| \x(0)) }_2^2 \right]. $$ One can demonstrate that the minimizer of that objective $\theta^*$ will be such that $s_{\theta^*}(\x,t) = \nabla_{\x} \log p_t(\x)$ \citep{vincent2011connection}, allowing us to approximate the reverse diffusion process. As can be seen, evaluating the objective requires the ability to generate samples from the FDP at arbitrary times $t$. Thankfully, as long as the drift is affine (i.e., $f(\x,t)=\A \x + \B$
), the transition kernel $p(\x(t)|\x(0))$ will always be normally distributed \citep{sarkka2019applied}, which means that we can solve the forward diffusion in a single step. Furthermore, the score of the Gaussian transition kernel is trivial to compute, making the loss an inexpensive training objective. 

There are two primary choices for the FDP in the literature, which we discuss below.

\subsection{Variance Exploding (VE) process}

The Variance Exploding (VE) process consists in the following FDP:
\begin{equation*}
    \dr \x = \sqrt{\frac{\dr \left[\sigma^2(t) \right]}{\dr t}}\dr \w.
\end{equation*}

Its associated transition kernel is:
\begin{equation*}
    \x(t)|\x(0) \sim \mathcal{N}(\x(0), [\sigma^2(t) - \sigma^2(0)] \I ) \approx \mathcal{N}(\x(0), \sigma^2(t)\I ).
\end{equation*}

In practice, we let $\sigma(t) =\sigma_{min}\left(\frac{\sigma_{max}}{\sigma_{min}}\right)^t$, where $\sigma_{min} = 0.01$ and $\sigma_{max}
\approx \max_{i} \sum_{j=1}^N ||\x^{(i)} - \
\x^{(j)} ||$ is the maximum Euclidean distance between two samples from the dataset $\{\x^{(i)}\}_{i=1}^N$ \citep{song2020improved}. Using the maximum Euclidean distance ensures that $\x(1)$ does not depend on $\x(0)$; thus, $\x(1)$ is approximately distributed as $\mathcal{N}(\Zero, 
\sigma^2(1)\I)$.

\subsection{Variance Preserving (VP) process}

The Variance Preserving (VP) process consists in the following FDP:
\begin{equation*}
    \dr \x = - \frac{1}{2} \beta(t) \x \dr t + \sqrt{\beta(t)} \dr\w.
\end{equation*}

Its associated transition kernel is:
\begin{equation*}
    \x(t)|\x(0) \sim \mathcal{N}(\x(0)~e^{-\frac{1}{2} \int_{0}^t \beta(s) \dr s}, (1- e^{-\int_{0}^t \beta(s) \dr s})~ \I ).
\end{equation*}

In practice, we let $\beta(t) = \beta_{min} + t \left( \beta_{max} - \beta_{min} \right)$, where $\beta_{min} = 0.1$ and $\beta_{max}=20$.
Thus, $\x(1)$ is approximately distributed as $\mathcal{N}(\Zero, \I)$ and does not depend on $\x(0)$.

\subsection{Solving the Reverse Diffusion Process (RDP)}

There are many ways to solve the RDP; the most basic one being Euler-Maruyama \citep{kloeden1992stochastic}, the SDE analog to Euler's method for solving ODEs.
\citet{song2020score} also proposed {\em Reverse-Diffusion}, which consists in ancestral sampling \citep{ho2020denoising} with the same discretization used in the FDP. With the Reverse-Diffusion, \citep{song2020score} obtained poor results unless applying an additional Langevin dynamics step after each Reversion-Diffusion step. They named this approach Predictor-Corrector (PC) sampling, with the predictor corresponding to Reverse-Diffusion and the corrector to Langevin dynamics. 
Although using a corrector step leads to better samples, PC sampling is only heuristically motivated and the theoretical underpinnings are not yet understood.
Nevertheless, \citep{song2020score} report their best results (in terms of lowest Fréchet Inception Distance \citep{heusel2017gans}) using the Reverse-Diffusion with Langevin dynamics for VE models. For VP models, they obtain their best results using Euler-Maruyama.

\section{Efficient Method for Solving Reverse Diffusion Processes}

\subsection{Setting up the algorithm}

We start with a general algorithm for solving an SDE (similar to most ODE/SDE solvers). We choose the various options/hyper-parameters based on a mixture of theory and experiments; an ablation study of the different hyper-parameters can also be found in Appendix \ref{sec:modif}.

\subsubsection{Integration method}\label{sec:int}

Solving the RDP to generate data can take an undesirably long time. One would assume that solving SDEs with high-order methods would improve speed over Euler-Maruyama, just like high-order ODE solvers improve speed over Euler's method when solving ODEs. However, this is not always the case: while higher-order solvers may achieve lower discretization errors, they require more function evaluations, 
and the improved precision might not be worth the increased computation cost
\citep{lehn2002adaptive, lamba2003adaptive}. 

Our preliminary attempts at using SDE solvers with the {\em DifferentialEquations.jl} Julia package \citep{DifferentialEquations.jl-2017} confirmed that higher-order methods were significantly slower (6 to 8 times slower; see Appendix \ref{sec:old}). Lamba's algorithm \citep{lamba2003adaptive}, a low-order adaptive method, yielded the fastest results, thus motivating us to restrict our search to the space of low-order methods. Still, the resulting images were of lower quality.

To be able to dynamically adjust the step size over time, thereby gaining speed over a fixed-step size algorithm, two integration methods are employed. Traditionally, for ODEs, a lower-order ($\x'$) method is used conjointly with a higher-order ($\x''$) one. The local error $E(\x',\x'') = \x' - \x''$ 
can be used to determine how stable the lower-order method is at the current step size; the closer to zero, the more appropriate the step size is. From this information, we can dynamically adjust the step size and decide whether or not to accept the proposed sample of the lower-order method. Alternatively, one can select $\x''$ as the proposal, which we will refer to as \em extrapolating\em.


Rather than using Improved Euler \citep{suli2003introduction} as in \citet{lamba2003adaptive} or a high-order stochastic Runge-Kutta method \citep{rossler2010runge} as in \citet{rackauckas2017adaptive} (which did not work well in our preliminary attempts with the Julia package) we instead rely on the stochastic Improved Euler's method \citep{roberts2012modify} as our higher order method. This method only requires two score function evaluations and re-use the same score function evaluation used for EM, meaning that it is only twice as expensive as EM. Similarly to Lamba's algorithm, this method, albeit quick, lead to images of poor quality. However, by using extrapolation (taking $\x''$ instead of $\x'$ as our proposal), we were able to match and improve over
the baseline approach (EM). Thus, using the stochastic Improved Euler was the key to taking bigger steps without sacrificing precision. Note that Lamba's algorithm cannot use extrapolation due to its use of a non-stochastic ODE solver (Improved Euler).

\subsubsection{Tolerance}

In ODE/SDE solvers, the local error is divided by 
a \em tolerance \em term.
Traditionally, the mixed tolerance $\bdelta(\x'): \mathbb{R}^d \to \mathbb{R}^d$ is calculated as the maximum between the absolute and relative tolerance:
\begin{equation} 
\bdelta(\x') = \max(\epsilon_{abs}, \epsilon_{rel}|\x'|),
\end{equation}
where the absolute value $|\cdot|$ is applied element-wise.

The {\em DifferentialEquations.jl} Julia package instead calculates the  mixed tolerance 
through the maximum of the current and previous sample:
\begin{equation} \label{eq:mixedtol2}
\bdelta(\x', \x'_{prev}) = \max(\epsilon_{abs}, \epsilon_{rel}\max(|\x'|, |\x'_{prev}|)).\end{equation}

Having no trivial prior for which approach to use, we tried both and found the latter approach (Equation \ref{eq:mixedtol2}) to converge much faster for VE models (see Appendix \ref{sec:modif}).

Given our focus on image generation, we can set $\epsilon_{abs}$ a priori. During training and at the end of the data generation, images are represented as floating-point tensors with range $[y_{min},y_{max}]$. When evaluated, they must be transformed into 8-bit color images; this means that images are scaled to the range $[0,255]$ and converted to the nearest integer (to represent one of the 256 values per color channel). Given the 8-bit color encoding, an absolute tolerance $\epsilon_{abs} = \frac{y_{max}-y_{min}}{256}$ corresponds to tolerating local errors of at most one color (e.g., $x_{ij}'$ with Red=5 and $x_{ij}''$ with Red=6 is accepted, but $x_{ij}'$ with Red=5 and $x_{ij}''$ with Red=7 is not) channel-wise. One-color differences are not perceptible and should 
not influence the metrics used for evaluating the generated images.
For VP models, which have range $[-1,1]$, this corresponds to $\epsilon_{abs} = 0.0078$ while for VE models, which have range $[0,1]$, this corresponds to $\epsilon_{abs} = 0.0039$.

To control speed/quality, we vary $\epsilon_{rel}$, where large values lead to more speed but less precision, while small values lead to the converse.

\subsubsection{Norm of the scaled error}

The scaled error (the error scaled by the mixed tolerance) is calculated as \begin{equation*} 
E_{q} = \norm{\frac{\x' - \x''}{\bdelta(\x', \x'_{prev})}}_q.
\end{equation*}

Many algorithms use $q=\infty$ \citep{lamba2003adaptive, rackauckas2017adaptive}, where $||\x||_{\infty} = \max(\x_1, ... , \x_k)$ over all $k$ elements of $\x$. Although this can work with low-dimensional SDEs, this is highly problematic for high-dimensional SDEs such as those in image-space. The reason is that a single channel of a single pixel (out of $65536$ pixels for a $256\times256$ color image) with a large local error will cause the step size to be reduced for all pixels and possibly lead to a step size rejection. Indeed, our experiments confirmed that using $q=\infty$ slows down generation considerably (see Appendix \ref{sec:modif}). To 
that effect, we instead use a scaled $\ell_2$ norm: 
\begin{equation*} ||\x||_{2} = \sqrt{\frac{1}{n} \sum_{i=1}^k (E_{q})_k}.\end{equation*}

\subsubsection{Hyperparameters of the dynamic step size algorithm}

Upon calculating the scaled error, we accept the proposal $\x''$ if $E_{q} \leq 1$
and 
increment the time by $h$
Whether or not it is accepted, we update the next step size $h$ in the usual way:
\begin{equation*}
    h \gets \min(h_{\max}, \theta h E_{q}^{-r}),
\end{equation*}
where $h_{\max}$ is the maximum step size, $\theta$ is the safety parameter which determines how strongly we adapt the step size (0 being very safe; 1 being fast, but high rejections rate), and $q$ is an exponent-scaling term.

Although ODE theory tells us that we should let $r=\frac{1}{p+1}$ with $p$ being the order of the lower-order integration method, there is no such theory for SDEs \citep{rackauckas2017adaptive}.
Thus, as \citet{rackauckas2017adaptive} suggest, we resorted to empirically testing values and found that any $r \in [0.5, 1]$ works well on both VE and VP processes, but that $r \in [0.8,0.9]$ is slightly faster (see Appendix \ref{sec:modif}). We arbitrarily chose $r=0.9$ as the default setting.

Finally, we defaulted to setting $\theta=0.9$ for the safety parameter as is common in the literature, and choose $h_{\max}$ to be equal to the largest step size possible, namely the remaining time $t$.

\subsubsection{Handling the mini-batch}

Using the same step size for every sample of a mini-batch means that every images would be slowed down by the other images. 
Since every image's RDP is independent from one another, we apply a different step size to each data sample; some images may converge faster than others, but we wait for all images to have converged. 

\subsection{Algorithm}

We have now defined every aspect of the algorithm needed to numerically solve the Equation (\ref{eq:backward}) for images. The resulting algorithm is described in Algorithm \ref{alg:anneal1}. This algorithm is straightforward to parallelize across the batch dimension.
Note that this algorithm is only for solving an RDP; a more general version for solving an arbitrary forward-time diffusion process can be found in Appendix \ref{sec:myalg}. Additionally, we present a demonstration in Appendix \ref{sec:stability} showing that the extrapolation step conserves the stability and convergence of the EM step.

\begin{algorithm}
	\caption{Dynamic step size extrapolation for solving Reverse Diffusion Processes}
	\label{alg:anneal1}
	\begin{algorithmic}
	    \Require{$s_\theta, \epsilon_{rel}, \epsilon_{abs}, h_{init}=0.01, r=0.9, \theta=0.9$} \algorithmiccomment{for images: $\epsilon_{abs} = \frac{y_{max}-y_{min}}{256}$}
        \State{$t \gets 1$}
        \State{$h \gets h_{init}$}
        \State{Initialize $\x$}
        \State{$\x_{prev}' \gets \x$}
        \While{$t > 0$}
            \State{Draw $\z \sim \mathcal{N}(\Zero, \I)$}
            \State{$\xp \gets \x - h f(\x,t) + h g(t)^2 s_{\theta}(\x,t) + \sqrt{h}g(t)\z$} \algorithmiccomment{Euler-Maruyama}
            \State{$\xheun \gets \x - h f(\xp,t- h) + h g(t- h)^2 s_{\theta}(\xp,t- h) + \sqrt{h}g(t - h)\z$}
            \State{$\xpp \gets \frac{1}{2}(\xp +\xheun)$} \algorithmiccomment{Improved Euler (SDE version)}
            \State{$\bdelta \gets \max(\epsilon_{abs}, \epsilon_{rel}\max(|\xp|, |\xp_{prev}|))$} \algorithmiccomment{Element-wise operations}
            \State{$ E_{2} \gets \frac{1}{\sqrt{n}}\norm{\left({\xp-\xpp}\right)/{\bdelta}}_2$}
            \If{$E_{2} \leq 1$} \algorithmiccomment{Accept}
                \State{$\x \gets \xpp$} \algorithmiccomment{Extrapolation}
                \State{$t \gets t - h$}
                \State{$\xp_{prev} \gets \xp$}
            \EndIf
	        \State{$h \gets \min(t,~\theta h E_{2}^{-r})$ \algorithmiccomment{Dynamic step size update}}
        \EndWhile
        \Return{$\x$}
        
	\end{algorithmic}
\end{algorithm}

\section{Experiments}

To test Algorithm \ref{alg:anneal1} on RDPs, we apply it to various pre-trained models from \citet{song2020score}. To start, we generate low-resolution images (32x32) by testing the VP, VE, VP-deep, and VE-deep models on CIFAR-10 \citep{krizhevsky2009learning}. Then, we generate higher-resolutions images (256x256) by testing the VE models on LSUN-Church \citep{yu2015lsun}, and Flickr-Faces-HQ (FFHQ) \citep{karras2019style}. 
See implementation details in Appendix \ref{sec:changes}.
We used four or less V100 GPUs to run the experiments.

To measure the performance of the image generation, we calculate the Fréchet Inception Distance (FID) \citep{heusel2017gans} and the Inception Score (IS) \citep{salimans2016improved}, where low FID and high IS correspond to higher quality/diversity. 
We compare our method to the three base solvers used in \citet{song2020score}: Reverse-Diffusion with Langevin dynamics, Euler-Maruyama (EM), and Probability Flow, where the latter solves an ODE instead of an SDE using Runge-Kutta 45 \citep{dormand1980family}. We also compare against the fast solver by \citep{song2020denoising} called denoising diffusion implicit models (DDIM), which is only defined for VP models. We define the {\em baseline} approach as the solver used by \citet{song2020score} which leads to the lowest FID (EM for VP models and Reverse-Diffusion with Langevin for VE models). For our algorithm, the only free hyperparameter is the relative tolerance which we set to $\epsilon_{rel} \in \{0.01, 0.02, 0.05, 0.1, 0.5\}$. 

The FID and the Number of score Function Evaluations (NFE) are described in Table \ref{tab:table1} for low-resolution images and Table \ref{tab:table2} for high-resolution images. The Inception Score (IS) is described for CIFAR-10 in Appendix~\ref{sec:inception}. 

\begin{table}
	\caption{Number of score Function Evaluations (NFE) / Fréchet Inception Distance (FID) on CIFAR-10 (32x32) from 50K samples}
	\label{tab:table1}
	\centering
	\begin{tabular}{ccccc}
		\toprule
		Method & VP & VP-deep & VE & VE-deep \\
		\cmidrule(){1-5}
		Reverse-Diffusion \& Langevin  & 1999 / 4.27 & 1999 / 4.69 &  1999 / {\fontseries{b}\selectfont 2.40} &  1999 / {\fontseries{b}\selectfont 2.21} \\
		\cmidrule(){1-5}
		Euler-Maruyama & 1000 / {\fontseries{b}\selectfont 2.55} & 1000 / {\fontseries{b}\selectfont 2.49} &  1000 / 2.98 &  1000 / 3.14 \\
		DDIM & 1000 / 2.86 & 1000 / 2.69 &  -- &  -- \\
		\cmidrule(){1-5}
		Ours ($\epsilon_{rel}=0.01$) & 329 / 2.70 & 330 / 2.56 & 738 / 2.91 & 736 / 3.06 \\
		Euler-Maruyama (same NFE) & 329 / 10.28 & 330 / 10.00 & 738 / 2.99 & 736 / 3.17 \\
		DDIM (same NFE) & 329 / 4.81 & 330 / 4.76 &  -- &  -- \\
		\cmidrule(){1-5}
		Ours ($\epsilon_{rel}=0.02$) & 274 / 2.74 & 274 / 2.60 & 490 / {\fontseries{b}\selectfont 2.87} & 488 / {\fontseries{b}\selectfont 2.99} \\
		Euler-Maruyama (same NFE) & 274 / 14.18 & 274 / 13.67 & 490 / 3.05 & 488 / 3.21 \\
		DDIM (same NFE) & 274 / 5.75 & 274 / 5.74 &  -- &  -- \\
		\cmidrule(){1-5}
		Ours ($\epsilon_{rel}=0.05$) & 179 / {\fontseries{b}\selectfont 2.59} & 180 / {\fontseries{b}\selectfont 2.44} & 271 / 3.23 & 270 / 3.40 \\
		Euler-Maruyama (same NFE) & 179 / 25.49 & 180 / 25.05 & 271 / 3.48 & 270 / 3.76 \\
		DDIM (same NFE) & 179 / 9.20 & 180 / 9.25 &  -- &  -- \\
		\cmidrule(){1-5}
		Ours ($\epsilon_{rel}=0.10$) & 147 / 2.95 & 151 / 2.73 & 170 / 8.85 & 170 / 10.15 \\
		Euler-Maruyama (same NFE) & 147 / 31.38 & 151 / 31.93 & 170 / 5.12 & 170 / 5.56 \\
		DDIM (same NFE) & 147 / 11.53 & 151 / 11.38 &  -- &  -- \\
		\cmidrule(){1-5}
		Ours ($\epsilon_{rel}=0.50$) & 49 / 72.29 & 48 / 82.42 & 52 / 266.75 & 50 / 307.32 \\
		Euler-Maruyama (same NFE)  & 49 / 92.99 & 48 / 95.77 & 52 / 169.32 & 50 / 271.27 \\
		DDIM (same NFE) & 49 / 37.24 & 48 / 38.71 &  -- &  -- \\
		\cmidrule(){1-5}
		Probability Flow (ODE) & 142 / 3.11 & 145 / 2.86 & 183 / 7.64 & 181 / 5.53 \\
		\bottomrule
	\end{tabular}
\end{table}

\begin{table}
	\caption{Number of score Function Evaluations (NFE) / Fréchet Inception Distance (FID) on LSUN-Church (256x256) and FFHQ (256x256) from 5K samples}
	\label{tab:table2}
	\centering
	\begin{tabular}{ccc}
		\toprule
		Method & VE (Church) & VE (FFHQ) \\
		\cmidrule(){1-3}
		Reverse-Diffusion \& Langevin  & 3999 / 29.14 & 3999 / 16.42 \\
		\cmidrule(){1-3}
		Euler-Maruyama & 2000 / 42.11 & 2000 / 18.57 \\
		\cmidrule(){1-3}
		Ours ($\epsilon_{rel}=0.01$) & 1104 / {\fontseries{b}\selectfont 25.67} & 1020 /{\fontseries{b}\selectfont 15.68} \\
		Euler-Maruyama (same NFE) & 1104 / 68.24 & 1020 / 20.45 \\
		\cmidrule(){1-3}
		Ours ($\epsilon_{rel}=0.02$) & 1030 / {\fontseries{b}\selectfont 26.46} & 643 / {\fontseries{b}\selectfont 15.67} \\
		Euler-Maruyama (same NFE) & 1030 / 73.47 & 643 / 44.42 \\
		\cmidrule(){1-3}
		Ours ($\epsilon_{rel}=0.05$) & 648 / 28.47 & 336 / 18.07 \\
		Euler-Maruyama (same NFE) & 648 / 145.96 & 336 / 114.23 \\
		\cmidrule(){1-3}
		Ours ($\epsilon_{rel}=0.10$) & 201 / 45.92 & 198 / 24.02 \\
		Euler-Maruyama (same NFE) & 201 / 417.77 & 198 / 284.61 \\
		\cmidrule(){1-3}
		Probability Flow (ODE) & 434 / 214.47 & 369 / 135.50 \\
		\bottomrule
	\end{tabular}
\end{table}

\subsection{Performance}

Compared to EM, we observe that our method is immediately advantageous in terms of quality/diversity for high-resolution images, along with $2$ to $3\times$ speedups ($\epsilon_{rel}=0.02$). While this advantage becomes less obvious in terms of the FID on CIFAR-10, our method still offers $>5\times$ computational speedups at no apparent disadvantage ($\epsilon_{rel} \in \{0.02, 0.05\}$).

Reverse-Diffusion with Langevin achieves the lowest FID for VE models on CIFAR-10, though at the cost of a $4\times$ computational overhead over our method. Furthermore, their advantage vanishes for VP models and when generating high-resolution images.
 
We further compare our SDE solver to EM given the same computational budget and observe that our method is always immensely preferable in high-resolutions and for VP models. For VE models on CIFAR-10, we observe that our algorithm leads to a better FID 
as long as the NFE is sufficiently large (270).
Note that since our algorithm takes two score function evaluations per step, EM has the advantage of doing twice as many steps given the same NFE, which appears to be a factor more important than the order of the method at low budget in low-resolution VE. 
Nevertheless, comparing for equal number of iterative step, the results still point to our method being always preferable.
For high-resolution images, we see that EM cannot converge on moderate to small NFEs due to the high-dimensionality, making of our SDE solver the way to go.

Generally, we observe that the VE process cannot be solved as fast as the VP process; this is due to the enormous Gaussian noise in the VE process causing larger local errors. This reflects the issue mentioned in Section \ref{sec:int} regarding high-order SDE solvers not always being beneficial in terms of speed for SDEs with heavy Gaussian noise. In practice, for VE, the algorithm uses a small step size in the beginning to ensure high accuracy and eventually increases the step size as the noise becomes less considerable.

\subsection{Solving an ODE instead of an SDE} We see that our SDE solver generally does better than Probability Flow, especially in high-resolution, where we obtain greatly lower FIDs with a similar budget. Our algorithm attains the same NFE as Probability Flow when $\epsilon_{rel}=0.10$ for low-resolution images and when $ 0.05 < \epsilon_{rel} < 0.10$ for high-resolution images. For the same budget, Probability Flow has higher FID than our approach on all but low-resolution VE models. However, in that case, our algorithm achieves a much lower FID when $\epsilon_{rel} \leq 0.05$, albeit slower. In high-resolution, Probability Flow leads to very poor FIDs, suggesting no convergence.

\subsection{DDIM} Contrary to \citet{song2020denoising},the FID of DDIM worsens significantly when the NFE decreases. This could be due to differences between \citet{song2020score} continuous-time score-matching and the DDIM training procedure and architecture. Nevertheless, the FID increase engendered by a reduced budget is much less dramatic than for EM. As of note, DDIM succeeds in maintaining a lower FID than our solver at extremely small NFEs ($< 50$), albeit with poor performances.

\section{Limitations}

Although we tested our approach on a wide range of settings, we nevertheless only tested on continuous-time image generation models. We did so because solving the SDE requires continuous-time and the only such pre-trained models at time of publishing are the one by \citet{song2020score}. 
Future work should train continuous-time versions of models from different data types, model architectures, and the learned-variance approach by \citet{nichol2021improved}, which generates data inherently faster; our SDE solver could then be used on these models.

Although our approach removes step size and schedule tuning, we still need to choose a value of the relative tolerance, which indirectly affects the number of steps taken; one could theoretically tune this hyper-parameter to optimize a certain metric, going against the point of removing tuning. Still, letting $\epsilon_{rel}=0.01$ for precise results and $\epsilon_{rel}=0.05$ for fast results are reasonable choices, as all evidence points to the FID being stable w.r.t. $\epsilon_{rel}$.

\section{Conclusion}

We built an SDE solver that allows for generating images of comparable (or better) quality to Euler-Maruyama 
at a much faster speed. 
Our approach makes image generation with score-based models more accessible by shrinking the required computational budgets by a factor of 2 to 5$\times$, and presenting a sensible way of compromising quality for additional speed. Nevertheless, data generation remains slow (a few minutes) compared to other generative models, which can generate data in a single forward pass of a neural network.
Therefore, additional work would be needed to find ways to make these models fast enough to be more attractive and useful for real-time applications.

\section{Broader Impact}
Our work allows generating data from score-based generative models more quickly, taking the technology closer to real-time applications.
While generative models have many socially beneficial applications, they can also be used to maliciously deceive humans (e.g. deepfakes). Generative models also carry the risk of reproducing biases in existing datasets.

\bibliographystyle{unsrtnat}
\interlinepenalty=10000
\bibliography{paper}

\clearpage

\appendix
\section*{Appendices}

\section{DifferentialEquations.jl} \label{sec:old}

\begin{table}[!ht]
	\caption{Short experiments with various SDE solvers from {\em DifferentialEquations.jl} on the VP model with a small mini-batch.}
	\label{tab:tableold}
	\centering
	\begin{tabular}{cccc}
		\toprule
		Method & Strong-Order & Adaptive & Speed \\
		\cmidrule(){1-4}
		Euler-Maruyama (EM) & 0.5 & No & Baseline speed \\
		SOSRA \citep{rossler2010runge} & 1.5 & Yes & {\color{red}5.92 times \textbf{slower}} \\
		SRA3 \citep{rossler2010runge} & 1.5 & Yes & {\color{red}6.93 times \textbf{slower}} \\
		Lamba EM (default) \citep{lamba2003adaptive} & 0.5 & Yes &  Did not converge \\
		Lamba EM (atol=1e-3) \citep{lamba2003adaptive} & 0.5 & Yes  & {\color{ForestGreen}2 times \textbf{faster}} \\
		Lamba EM (atol=1e-3, rtol=1e-3) \citep{lamba2003adaptive} & 0.5 & Yes  & {\color{ForestGreen}1.27 times \textbf{faster}} \\
		Euler-Heun & 0.5 & No  & {\color{red}1.86 times \textbf{slower}} \\
		Lamba Euler-Heun \citep{lamba2003adaptive} & 0.5 & Yes &  {\color{ForestGreen}1.75 times \textbf{faster}} \\
		SOSRI \citep{rossler2010runge} & 1.5 & Yes &  {\color{red}8.57 times \textbf{slower}} \\
		RKMil (at various tolerances) \citep{kloeden1992stochastic} & 1.0 & Yes & Did not converge \\
		ImplicitRKMil \citep{kloeden1992stochastic} & 1.0 & Yes & Did not converge \\
		ISSEM & 0.5 & Yes & Did not converge \\
		\bottomrule
	\end{tabular}
\end{table}

Here, we report the preliminary experiments we ran with the {\em DifferentialEquations.jl} Julia package \citep{DifferentialEquations.jl-2017} before devising our own SDE solver. As can be seen, most methods either did not converge (with warnings of "instability detected") or converged, but were much slower than Euler-Maruyama. The only promising method was Lamba's method \citep{lamba2003adaptive}. Note that an algorithm has strong-order $p$ when the local error from $t$ to $t+h$ is $\mathcal{O}(h^{p+1})$).

\section{Effects of modifying Algorithm \ref{alg:anneal1}}\label{sec:modif}

\begin{table}[!ht]
	\caption{Effect of different settings on the [Inception score (IS) / Fréchet Inception Distance (FID) / Number of score Function Evaluations (NFE)] from 10k samples (with mini-batches of 1k samples) with the VP - CIFAR10 model.}
	\label{tab:table2a}
	\centering
	\begin{tabular}{lccc}
		\toprule
		Change(s) in Algorithm \ref{alg:anneal1} & IS & FID & NFE \\
		\cmidrule(){1-4}
	    No change $\left[ q=2, r=0.9, \delta(\x',\x'_{prev})\right]$ & 9.38 & 4.70 & 3972 \\
	    \cmidrule(){1-4}
	    \multicolumn{4}{c}{Small modifications} \\
	    \cmidrule(){1-4}
	   $\delta(\x')$ & 9.26 & 4.69 & 4166 \\
	    No Extrapolation (thus, using Euler–Maruyama) & 9.58 & 11.73 & 3978 \\
	    $q=\infty$ & 9.48 & 4.90 & 14462 \\
	    $r=.5$ & 9.41 & 4.69 & 4104 \\
	    $r=.8$ & 9.36 & 4.68 & 3938 \\
	    $r=1$  & 9.41 & 4.69 & 4048 \\
	    \cmidrule(){1-4}
	    \multicolumn{4}{c}{Variations of \citet{lamba2003adaptive} Algorithm}\\
	    \cmidrule(){1-4}
	    $r=0.5$, Lamba integration & 7.80 & 52.98 & 1468 \\
	    $r=0.5$, Lamba integration, Extrapolation & 7.32 & 64.65 & 1438 \\
	   $r=0.5$, Lamba integration, $q=\infty$ & 9.28 & 21.09 & 2360 \\
	   $r=0.5$, Lamba integration, $q=\infty$, $\theta=0.8$ & 9.21 & 18.82 & 2346 \\
		\bottomrule
	\end{tabular}
\end{table}

\begin{table}[!ht]
	\caption{Effect of different settings on the [Inception score (IS) / Fréchet Inception Distance (FID) / Number of score Function Evaluations (NFE)] from 10k samples (with mini-batches of 1k samples) with the VE - CIFAR10 model.}
	\label{tab:table2b}
	\centering
	\begin{tabular}{lccc}
		\toprule
		Change(s) in Algorithm \ref{alg:anneal1} & IS & FID & NFE \\
		\cmidrule(){1-4}
	    No change $\left[ q=2, r=0.9, \delta(\x',\x'_{prev})\right]$ & 9.39 & 4.89 & 8856 \\
	    \cmidrule(){1-4}
	    \multicolumn{4}{c}{Small modifications} \\
	    \cmidrule(){1-4}
	   $\delta(\x')$ & 9.39 & 4.99 & 17514 \\
	    No Extrapolation (thus, using Euler–Maruyama) & 9.58 & 6.57 & 8802 \\
	    $q=\infty$ & 9.41 & 5.03 & 39500 \\
	    $r=0.5$ & 9.47 & 4.87 & 9594 \\
	    $r=0.8$ & 9.45 & 4.84 & 8952 \\
	    $r=1$  & 9.43 & 4.93 & 8784 \\
	    \cmidrule(){1-4}
	    \multicolumn{4}{c}{Variations of \citet{lamba2003adaptive} Algorithm}\\
	    \cmidrule(){1-4}
	   $r=0.5$, Lamba integration & 9.08 & 18.28 & 2492 \\
	   $r=0.5$, Lamba integration, Extrapolation & 3.70 & 169.78 & 2252 \\
	   $r=0.5$, Lamba integration, $q=\infty$ & 9.42 & 6.80 & 5886 \\
	   $r=0.5$, Lamba integration, $q=\infty$, $\theta=0.8$ & 9.35 & 6.20 & 2970 \\
		\bottomrule
	\end{tabular}
\end{table}

As can be seen, most chosen settings are lead to better results. However, $r$ seems to have little impact on the FID. Still, using $r \in [0.8, 0.9]$ lead to a little bit less score function evaluations and sometimes lead to lower FID.

\clearpage

\section{Dynamic step size algorithm for solving any type of SDE (rather than just Reverse Diffusion Processes)}\label{sec:myalg}

Assume, we have a Diffusion Process of the form: \begin{equation}\label{eq:forward2}
\dr\x = f(\x,t)\dr t + g(\x,t)\dr \w.
\end{equation}
The algorithm to solve it is represented in Algorithm \ref{alg:anneal2}. The differences are:
\begin{itemize}
    \item it is in forward-time
    \item the range of time must be given
    \item The diffusion can depend on $\x$, which leads to a slightly more complicated formulation that depends on some random number $s = \pm 1$ \citep{roberts2012modify}.
    \item we retain the full trajectory instead of only the ending
    \item we retain the noise after a rejection to ensure that there is no bias in the rejections
\end{itemize}

\begin{algorithm}[H]
	\caption{Dynamic step size extrapolation for solving arbitrary (forward-time) Diffusion Processes}
	\label{alg:anneal2}
	\begin{algorithmic}
	    \Require{$s_\theta, t_{begin}, t_{end}, \epsilon_{rel}, \epsilon_{abs}, h_{init}=0.01, r=0.9, \theta=0.9$}
        \State{$t \gets t_{begin}$}
        \State{$h \gets h_{init}$}
        \State{Initialize $\x(t)$}
        \State{$\x_{prev}' \gets \x$}
        \State{Draw $\z \sim \mathcal{N}(\Zero, \I)$}
        \While{$t < t_{end}$}
            \If{Stratonovich SDE or $g(\x,t)=g(\x)$}
                \State{$s \gets 0$}
            \Else{} \algorithmiccomment{Itō diffusion}
                \State{Draw $s \sim \mathrm{Uniform}(\{-1, 1\})$}
            \EndIf
            \State{$\xp \gets \x(t) + h f(\x(t),t) + \sqrt{h}g(\x(t), t)(\z - s)$} \algorithmiccomment{Euler-Maruyama}
            \State{$\xheun \gets \x(t) + h f(\xp,t+ h) + \sqrt{h}g(\x',t+ h)(\z + s)$}
            \State{$\xpp \gets \frac{1}{2}(\xp +\xheun)$} \algorithmiccomment{Improved Euler (SDE version)}
            \State{$\bdelta \gets \max(\epsilon_{abs}, \epsilon_{rel}\max(|\xp|, |\xp_{prev}|))$} \algorithmiccomment{Element-wise operations}
            \State{$ E_{2} \gets \frac{1}{\sqrt{n}}\norm{\left({\xp-\xpp}\right)/{\bdelta}}_2$}
            \If{$E_{2} \leq 1$} \algorithmiccomment{Accept}
                \State{$t \gets t + h$}
                \State{$\x(t) \gets \xpp$} \algorithmiccomment{Extrapolation}
                \State{$\xp_{prev} \gets \xp$}
                \State{Draw $\z \sim \mathcal{N}(\Zero, \I)$}
            \EndIf
	        \State{$h \gets \min(t,~\theta h E_{2}^{-r})$ \algorithmiccomment{Dynamic step size update}}
        \EndWhile
        \Return{$\x$}
        
	\end{algorithmic}
\end{algorithm}

\clearpage

\section{Implementation Details} \label{sec:changes}

We started from the original code by \citet{song2020score} but changed a few settings concerning the SDE solving. This create some very minor difference between their reported results and ours. For the VP and VP-deep models, we obtained 2.55 and 2.49 instead of the original 2.55 and 2.41 for the baseline method (EM). For the VE and VE-deep models, we obtained 2.40 and 2.21 instead of the original 2.38 and 2.20 for the baseline method (Reverse-Diffusion with Langevin).

When solving the SDE, time followed the sequence $t_0 = 1$, $t_i = t_{i-1} - \frac{1-\epsilon}{N}$, where $N=1000$ for CIFAR-10, $N=2000$ for LSUN, $\epsilon = 1e-3$ for VP models, and $\epsilon = 1e-5$ for VE models.

Meanwhile, the actual step size $h$ used in the code for Euler-Maruyama (EM) was equal to $\frac{1}{N}$. Thus, there was a negligible difference between the step size used in the algorithm ($h=\frac{1}{N}$) and the actual step size implied by $t$ ($h=\frac{1-\epsilon}{N}$). Note that this has little to no impact.

The bigger issue is at the last predictor step was going from $t=\epsilon$ to $t=\epsilon - \frac{1}{N} < 0$. Thus, $t$ was made negative. Furthermore the sample was denoised at $t < 0$ while assuming $t=\epsilon$. There are two ways to fix this issue: 1) take only a step from $t=\epsilon$ to $t=0$ and do not denoise (since you cannot denoise with the incorrect $t$ or with $t=0$), or 2) stop at $t=\epsilon$ and then denoise. Since denoising is very helpful, we took approach 2; however, both approaches are sensible.

Finally, denoising was not implemented correctly before. Denoising was implemented as one predictor step (Reverse-Diffusion or EM) without adding noise. This corresponds to: \begin{equation*}
    \x \gets \x - h \left[ f(\x,t)-g(t)^2\nabla_{\x} \log p_t(\x) \right].
    \end{equation*}
At the last iteration, this incorrect denoising would be:
\begin{align*} \x &\gets \x + \frac{\dr[\sigma^2(t)]}{\dr t} \frac{1}{N} \nabla_{\x} \log p_t(\x) \\ 
&= \x + \frac{\sigma_{min}}{N} \sqrt{2 \log\left(\frac{\sigma_{max}}{\sigma_{min}}\right)} \nabla_{\x} \log p_t(\x) \\
&\approx \x
\end{align*} for VE and 
\begin{align*}\x &\gets \x + \frac{\sqrt{\beta_{min}}}{N} \nabla_{\x} \log p_t(\x) \\
&\approx \x
\end{align*} for VP.

Meanwhile, the correct way to denoise based on Tweedie formula \citep{efron2011tweedie} is:
\begin{equation*}
 \x \gets \x + \text{Var}[\x(t)|\x(0)] \nabla_{\x} \log p_t(\x),
\end{equation*}
where $\text{Var}[\x(t)|\x(0)]$ is the variance of the transition kernel: $\text{Var}[\x(t)|\x(0)] = \sigma_{min}=0.01$ for VE and $\text{Var}[\x(t)|\x(0)] = 1$.
This means that the correct Tweedie formula corresponds to
\begin{align*} \x &\gets \x + 0.01^2 \nabla_{\x} \log p_t(\x) \\
&\approx \x
\end{align*} for VE and $$\x \gets \x + \nabla_{\x} \log p_t(\x)$$ for VP.

As can be seen, denoising has a very small impact on VE so the difference between the correct and incorrect denoising is minor. Meanwhile, for VP the incorrect denoising lead to a tiny change, while the correct denoising lead to a large change. In practice, we observe that changing the denoising method to the correct one does not significantly affect the FID with VE, but lowers down the FID significantly with VP.

\clearpage





\section{Inception Score on CIFAR-10}\label{sec:inception}

\begin{table}[!htbp]
	\caption{Inception Score on CIFAR-10 (32x32) from 50K samples}
	\label{tab:table3}
	\centering
	\begin{tabular}{ccccc}
		\toprule
		Method & VP & VP-deep & VE & VE-deep \\
		\cmidrule(){1-5}
		Reverse-Diffusion \& Langevin  & 9.94 & 9.85 &  9.86 &  9.83 \\
		\cmidrule(){1-5}
		Euler-Maruyama & 9.71 & 9.73 & 9.49 & 9.31 \\
		Ours ($\epsilon_{rel}=0.01$) & 9.46 & 9.54 & 9.50 & 9.48 \\
		Ours ($\epsilon_{rel}=0.02$) & 9.51 & 9.48 & 9.57 & 9.50 \\
		Ours ($\epsilon_{rel}=0.05$) & 9.50 & 9.61 & 9.64 & 9.63 \\
		Ours ($\epsilon_{rel}=0.10$) & 9.69 & 9.64 & 9.87 & 9.75 \\
		Probability Flow (ODE) & 9.37 & 9.33 & 9.17 & 9.32 \\
		\bottomrule
	\end{tabular}
\end{table}

\section{Stability and Bias of the Numerical Scheme}\label{sec:stability}
The following constructions rely on the underlying
assumption of the stochastic dynamics being driven by a wiener process.
More so, we also assume that the Brownian motion is
time symmetrical. Both assumptions are consistent and widely used
in the literature; for example, see \citep{gardiner2009stochastic}
\citep{arnold1974stochastic}.

The method described in Algorithm 1 gives us a significant speedup
in terms of computing time and actions. Albeit the speed up comes from
a piece-wise step in the algorithm combining the traditional Euler
Maruyama (EM) with a form of adaptive step size predictor-corrector.
Here we show that both the stability and the convergence of the EM
scheme are conserved by introducing the extra adaptive stepsize
of our new scheme. As a first step, we define the stability and bias
in a Stochastic Differential Equation (SDE) numerical solution. 

We denote $\Re(\lambda)$ as the real value of a complex-valued $\lambda$.

The linear test SDE is defined in the following way:
\begin{equation}
\dr\x_{t}=\lambda \x_{t}\dr t+\sigma \dr \w_{t}\label{eq:linear_test}
\end{equation}
with its numerical counterpart 
\[
\y_{n+1}=\Re\left(h\lambda\right)\y_{n}+\z_{n},
\]
 where the $\z_{n}$ are random variables that do not depend on $\y_{0},\y_{1}......\y_{n}$
or $\lambda$ and the EM scheme is 
\[
\y_{n+1}=\left(1+h\lambda\right)\y_{n}+\z_{n}.
\]

A numerical scheme is asymptotically unbiased with step size $h>0$
if, for a given linear SDE (\ref{eq:linear_test}) driven by a two-sided
Wiener process, the distribution of the numerical solution $\y_{n}$
converges as $n\rightarrow\infty$ to the normal distribution with
zero mean and variance $\frac{\sigma^{2}}{2\left|\lambda\right|}$\citep{artemiev2011numerical}.
This stems from the fact that a solution of a linear SDE (\ref{eq:linear_test})
is a Gaussian process whenever the initial condition is Gaussian (or
deterministic); thus, there are only two moments that control the
bias in the algorithm: 
\[
\lim_{n\rightarrow\infty}\mathbb{E}\left[\y_{n}\right]=0,
\hspace{0.8cm}\lim_{n\rightarrow\infty}\mathbb{E}\left[\y_{n}^{2}\right]=-\frac{\sigma^{2}}{2\left|\lambda\right|}.
\]

A numerical scheme with step size $h$ is numerically stable in mean
if the numerical solution $\y_{n}^{\left(h\right)}$ applied to a linear
SDE satisfies 
\[
\lim_{n\rightarrow\infty}\mathbb{E}\left[\y_{n}\right]=0,
\]
and is stable in mean square \citep{saito1996stability} if we have that
\[
\lim_{h\rightarrow0}\left(\lim_{n\rightarrow\infty}\mathbb{E}\left[\left|\y_{n}\right|^{2}\right]\right)=\frac{\sigma^{2}}{2\Re(\lambda)}.
\]

In what follows, we will trace the criteria for bias through our algorithm
and show that it remains unbiased. By construction, the first
EM step remains unbiased, while for the RDP, we write down the time
reverse Wiener process as
\[
\tilde{\y}_{n+1}=\left(1+\lambda h\right)\tilde{\y}_{n}+\tilde{\z}_{n}
\]
in the reverse time steps $h$ i.e., $t-nh$, $t-2nh$,
\begin{align*}
\mathbb{E}\left[\tilde{\y}_{n+1}\right] &=\left(1+\lambda\left(t-h\right)\right)\mathbb{E}\left[\tilde{\y}_{n}\right] \\ 
 &= \left(1+\lambda\left(t-h\right)\right)\mathbb{E}\left[\left(1+\lambda\left(t-h\right)\right)\tilde{\y}_{n-1}\right]\\
 & \hspace{2cm}\vdots \\
 &= \left(1+\lambda\left(t-h\right)\right)^{n+1}\mathbb{E}\left[\tilde{\y}_{0}\right]\\
 &= \left(1+\lambda\left(t-h\right)\right)^{n+1}\mathbb{E}\left[\y_{0}\right].
\end{align*}
Thus, if 
\[
\left|1+\lambda\left(t-h\right)\right|<1,
\]
then 
\[
\lim_{n\rightarrow\infty}\mathbb{E}\left[\y_{n}^{\left(h\right)}\right]=0.
\]
In Algorithm 1, we are performing consecutive steps forward and backwards
in time so $t=2h$ such that
\[
\left|1+\lambda h\right|<1.
\]
Thus, the scheme is both numerically stable and unbiased with respect to the mean. 

Next, we focus on the numerical solution in mean square:
\begin{align*}
\mathbb{E}\left[\left|\tilde{\y}_{n+1}\right|^{2}\right] & =\left|1+\lambda\left(t-h\right)\right|^{2}\mathbb{E}\left[\left|\tilde{\y}_{n}\right|^{2}\right]+\sigma^{2}h\\
 &= \left|1+\lambda\left(t-h\right)\right|^{2}\left\{ \left|1+\lambda\left(t-h\right)\right|^{2}\mathbb{E}\left[\left|\tilde{\y}_{n-1}\right|^{2}\right]+\sigma^{2}h\right\} +\sigma^{2}h\\
 & \hspace{2cm}\vdots \\
 & = \left|1+\lambda\left(t-h\right)\right|^{2\left(n+1\right)}\mathbb{E}\left[\left|\y_{0}\right|\right]+\frac{\left|1+\lambda\left(t-h\right)\right|^{2\left(n+1\right)}-1}{2\Re\lambda+\left|\lambda\right|^{2}\left(t-h\right)}\sigma^{2}.
\end{align*}
Under the same assumption of consecutive steps, we have that
\[
\mathbb{E}\left[\left|\tilde{\y}_{n+1}\right|^{2}\right]=\left|1+\lambda h\right|^{2\left(n+1\right)}\mathbb{E}\left[\left|\y_{0}\right|\right]+\frac{\left|1+\lambda h\right|^{2\left(n+1\right)}-1}{2\Re(\lambda)+\left|\lambda\right|^{2}h}\sigma^{2},
\]
\[
\lim_{n\rightarrow\infty}\mathbb{E}\left[\left|\tilde{\y}_{n+1}\right|^{2}\right]=-\frac{\sigma^{2}}{2\Re(\lambda)+\left|\lambda\right|^{2}h},
\]
\[
\lim_{h\rightarrow0}\left(\lim_{n\rightarrow\infty}\mathbb{E}\left[\left|\tilde{\y}_{n+1}\right|^{2}\right]\right)=-\frac{\sigma^{2}}{2\Re(\lambda)}.
\]
Assuming the imaginary part of $\lambda$ is null, we have that
\[
\lim_{h\rightarrow0}\left(\lim_{n\rightarrow\infty}\mathbb{E}\left[\left|\tilde{\y}_{n+1}\right|^{2}\right]\right)=-\frac{\sigma^{2}}{2\left|\lambda\right|}.
\]
Thus, the numerical scheme is stable and unbiased in the mean square.

Following the two steps for computation of $\x'$ and
$\xheun$, the step size decreases and does not
change size; thus, all the above statements hold, and the entire algorithm is stable and unbiased with respect to both the mean and square mean.

\clearpage
\section{Samples}

\begin{figure}[ht] 
  \begin{subfigure}[b]{0.5\linewidth}
    \centering
    \includegraphics[width=1\linewidth]{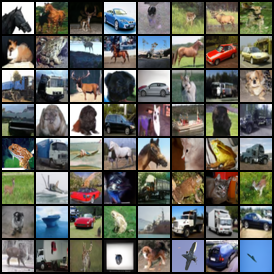} 
    \caption{Dynamic-step Extrapolation ($\epsilon=0.01$)} 
  \end{subfigure}
  \begin{subfigure}[b]{0.5\linewidth}
    \centering
    \includegraphics[width=1\linewidth]{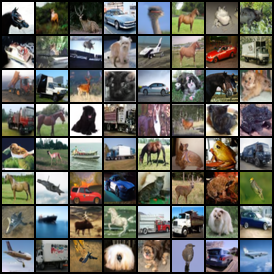}
    \caption{Dynamic-step Extrapolation ($\epsilon=0.02$)} 
  \end{subfigure}
  \begin{subfigure}[b]{0.5\linewidth}
    \centering
    \includegraphics[width=1\linewidth]{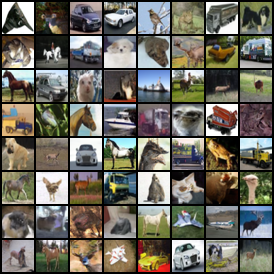} 
    \caption{Dynamic-step Extrapolation ($\epsilon=0.05$)} 
  \end{subfigure}
  \begin{subfigure}[b]{0.5\linewidth}
    \centering
    \includegraphics[width=1\linewidth]{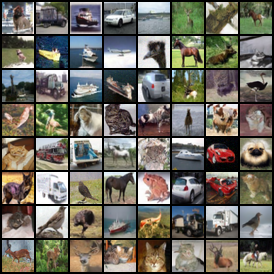}
    \caption{Dynamic-step Extrapolation ($\epsilon=0.10$)} 
  \end{subfigure}
  \caption{VP - CIFAR10}
  \label{fig11} 
\end{figure}

\begin{figure}[ht] 
  \begin{subfigure}[b]{0.5\linewidth}
    \centering
    \includegraphics[width=1\linewidth]{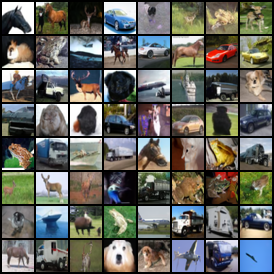} 
    \caption{Dynamic-step Extrapolation ($\epsilon=0.01$)} 
  \end{subfigure}
  \begin{subfigure}[b]{0.5\linewidth}
    \centering
    \includegraphics[width=1\linewidth]{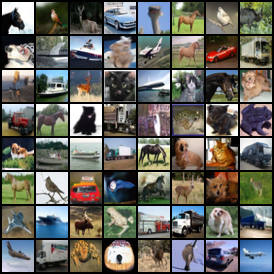}
    \caption{Dynamic-step Extrapolation ($\epsilon=0.02$)} 
  \end{subfigure}
  \begin{subfigure}[b]{0.5\linewidth}
    \centering
    \includegraphics[width=1\linewidth]{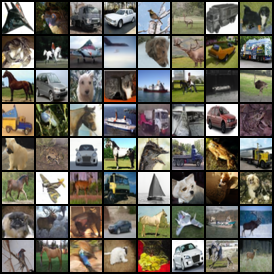} 
    \caption{Dynamic-step Extrapolation ($\epsilon=0.05$)} 
  \end{subfigure}
  \begin{subfigure}[b]{0.5\linewidth}
    \centering
    \includegraphics[width=1\linewidth]{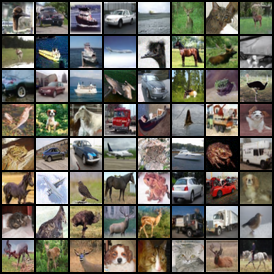}
    \caption{Dynamic-step Extrapolation ($\epsilon=0.10$)} 
  \end{subfigure}
  \caption{VP-deep - CIFAR10}
  \label{fig12} 
\end{figure}

\begin{figure}[ht] 
\begin{subfigure}[b]{0.5\linewidth}
    \centering
    \includegraphics[width=1\linewidth]{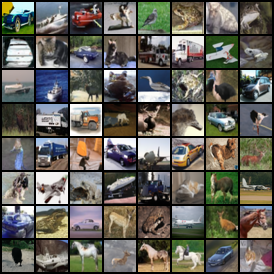} 
    \caption{Dynamic-step Extrapolation ($\epsilon=0.01$)} 
  \end{subfigure}
  \begin{subfigure}[b]{0.5\linewidth}
    \centering
    \includegraphics[width=1\linewidth]{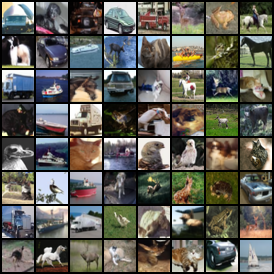}
    \caption{Dynamic-step Extrapolation ($\epsilon=0.02$)} 
  \end{subfigure}
  \begin{subfigure}[b]{0.5\linewidth}
    \centering
    \includegraphics[width=1\linewidth]{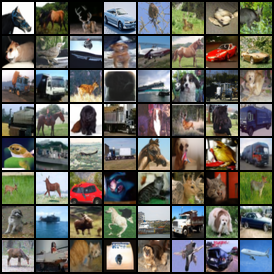} 
    \caption{Dynamic-step Extrapolation ($\epsilon=0.05$)} 
  \end{subfigure}
  \begin{subfigure}[b]{0.5\linewidth}
    \centering
    \includegraphics[width=1\linewidth]{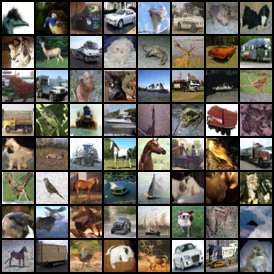}
    \caption{Dynamic-step Extrapolation ($\epsilon=0.10$)} 
  \end{subfigure}
  \caption{VE - CIFAR10}
  \label{fig13} 
\end{figure}

\begin{figure}[ht] 
  \begin{subfigure}[b]{0.5\linewidth}
    \centering
    \includegraphics[width=1\linewidth]{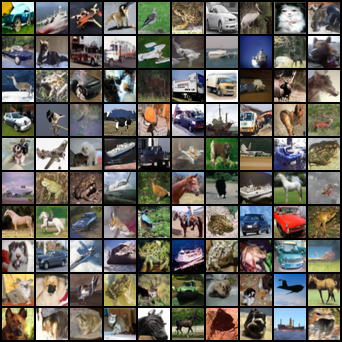} 
    \caption{Dynamic-step Extrapolation ($\epsilon=0.01$)} 
  \end{subfigure}
  \begin{subfigure}[b]{0.5\linewidth}
    \centering
    \includegraphics[width=1\linewidth]{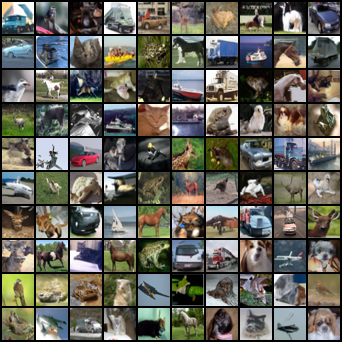}
    \caption{Dynamic-step Extrapolation ($\epsilon=0.02$)} 
  \end{subfigure}
  \begin{subfigure}[b]{0.5\linewidth}
    \centering
    \includegraphics[width=1\linewidth]{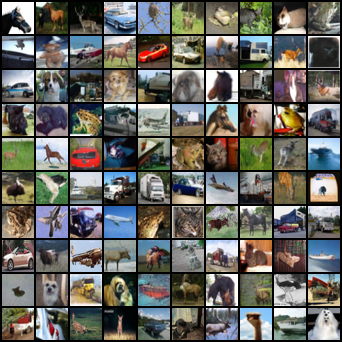} 
    \caption{Dynamic-step Extrapolation ($\epsilon=0.05$)} 
  \end{subfigure}
  \begin{subfigure}[b]{0.5\linewidth}
    \centering
    \includegraphics[width=1\linewidth]{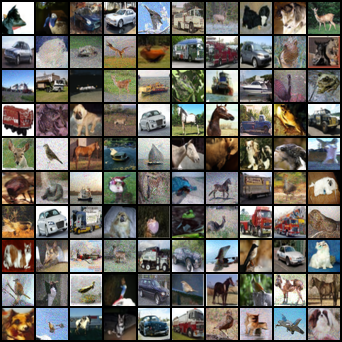}
    \caption{Dynamic-step Extrapolation ($\epsilon=0.10$)} 
  \end{subfigure}
  \caption{VE-deep - CIFAR10}
  \label{fig14} 
\end{figure}

\begin{figure}[ht] 
  \begin{subfigure}[b]{0.5\linewidth}
    \centering
    \includegraphics[width=1\linewidth]{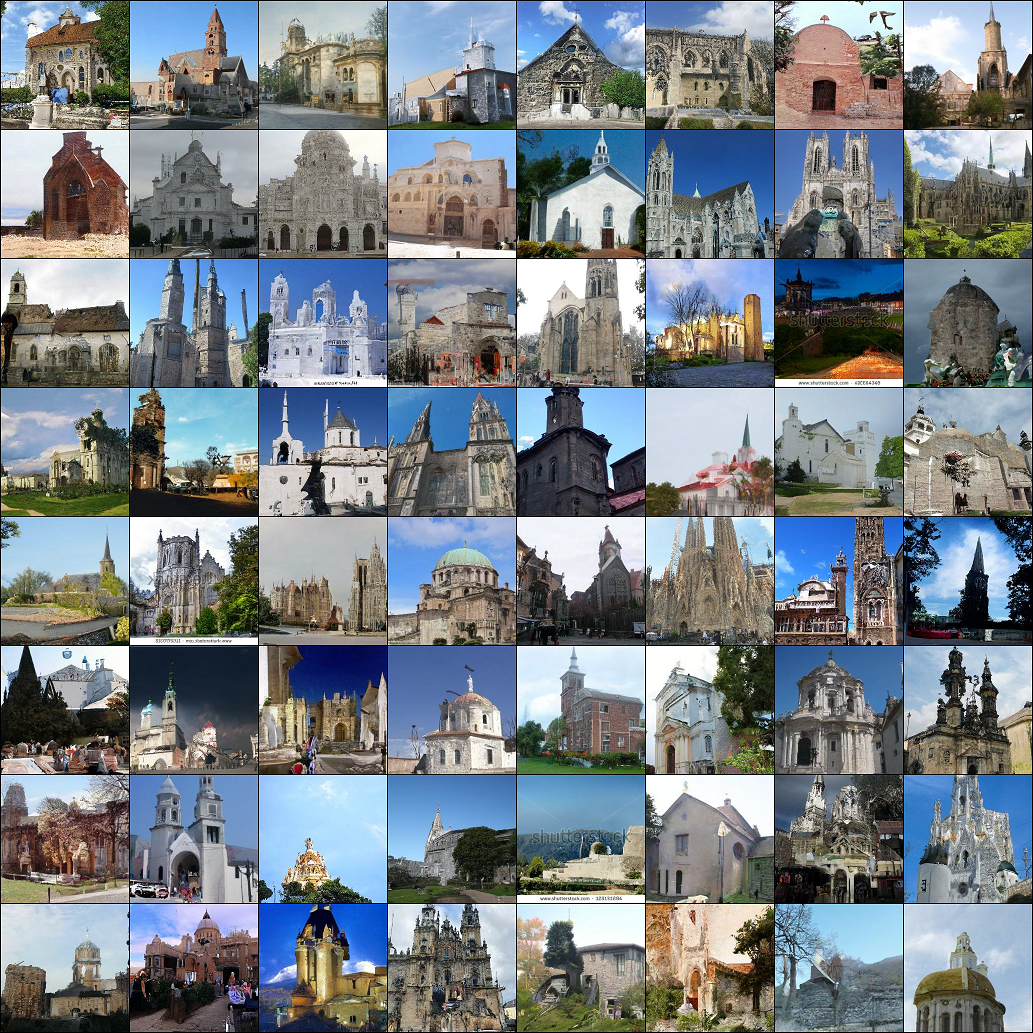} 
    \caption{Dynamic-step Extrapolation ($\epsilon=0.01$)} 
  \end{subfigure}
  \begin{subfigure}[b]{0.5\linewidth}
    \centering
    \includegraphics[width=1\linewidth]{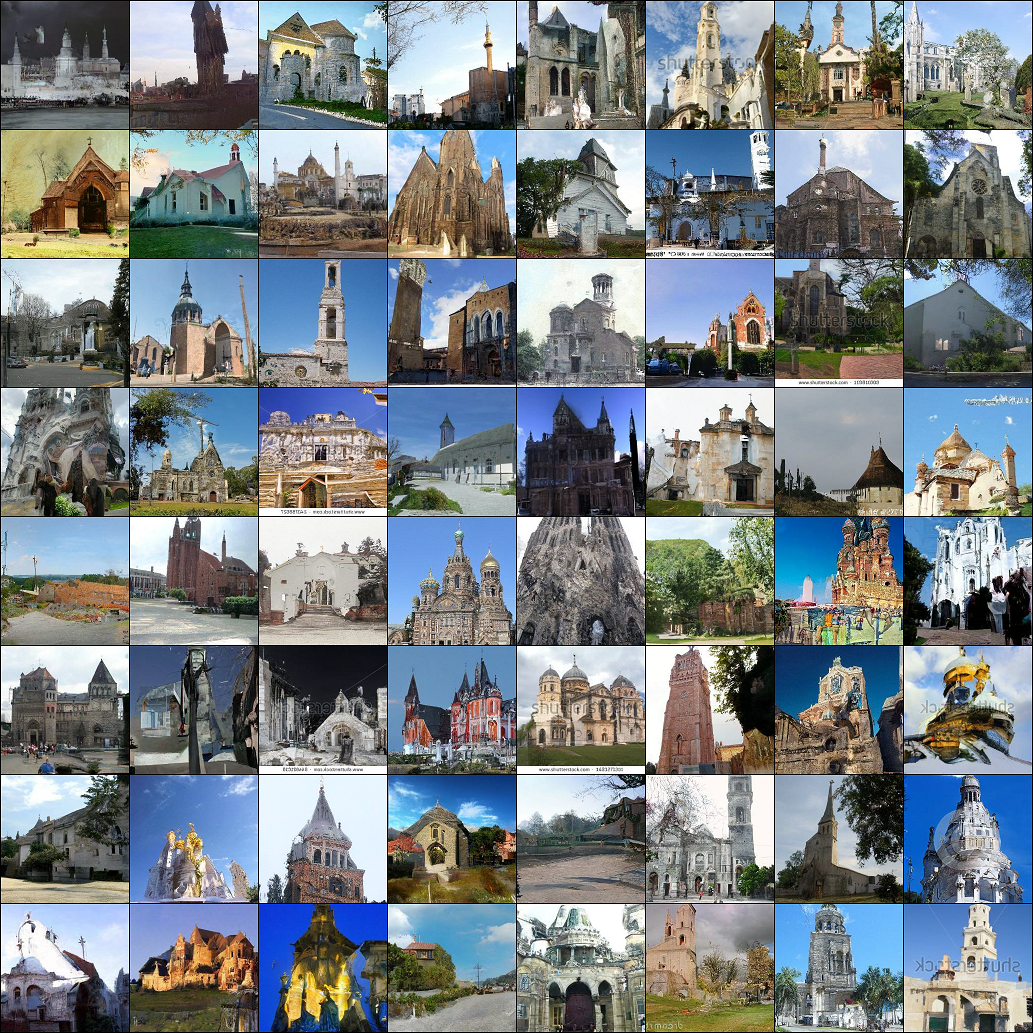}
    \caption{Dynamic-step Extrapolation ($\epsilon=0.02$)} 
  \end{subfigure}
  \begin{subfigure}[b]{0.5\linewidth}
    \centering
    \includegraphics[width=1\linewidth]{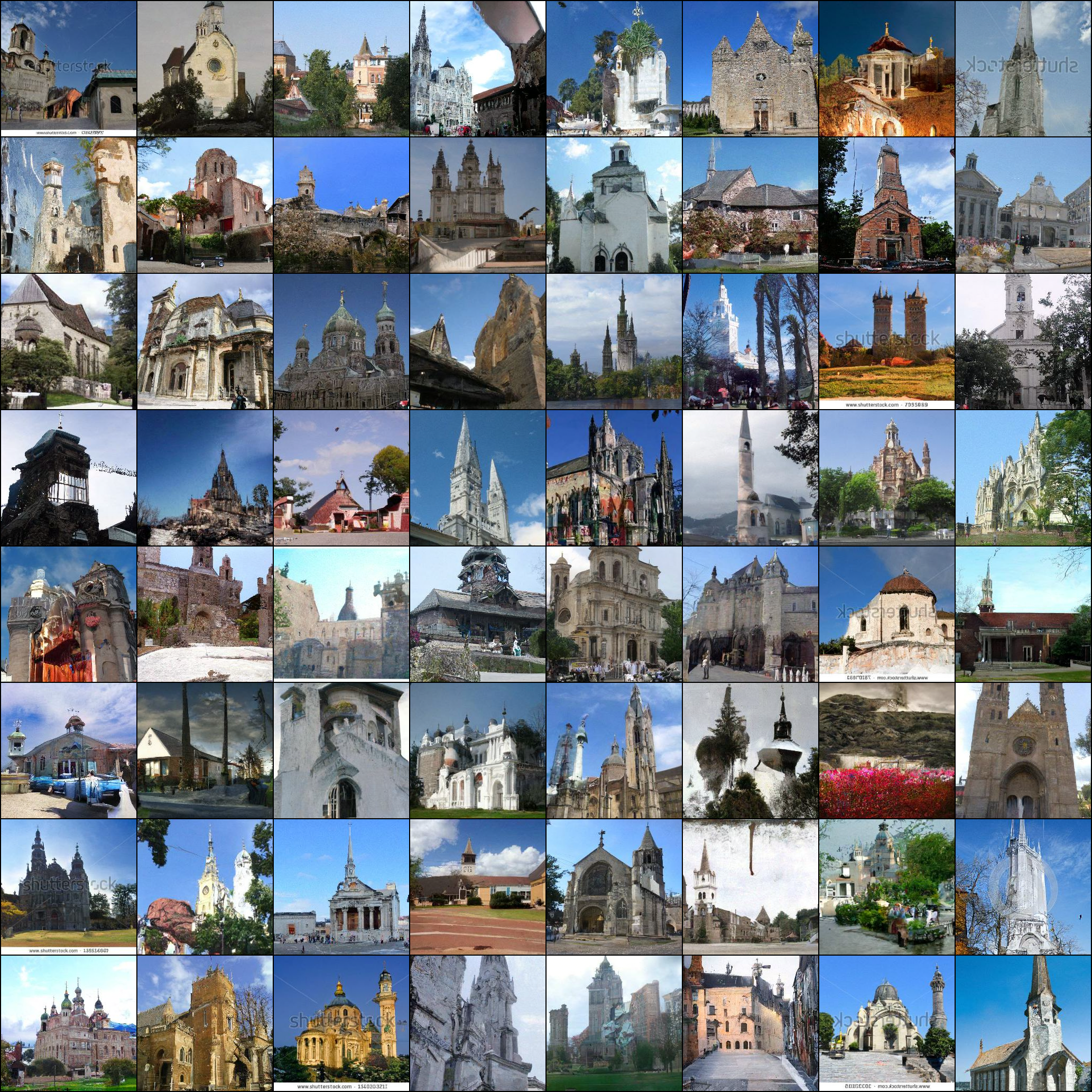} 
    \caption{Dynamic-step Extrapolation ($\epsilon=0.05$)} 
  \end{subfigure}
  \begin{subfigure}[b]{0.5\linewidth}
    \centering
    \includegraphics[width=1\linewidth]{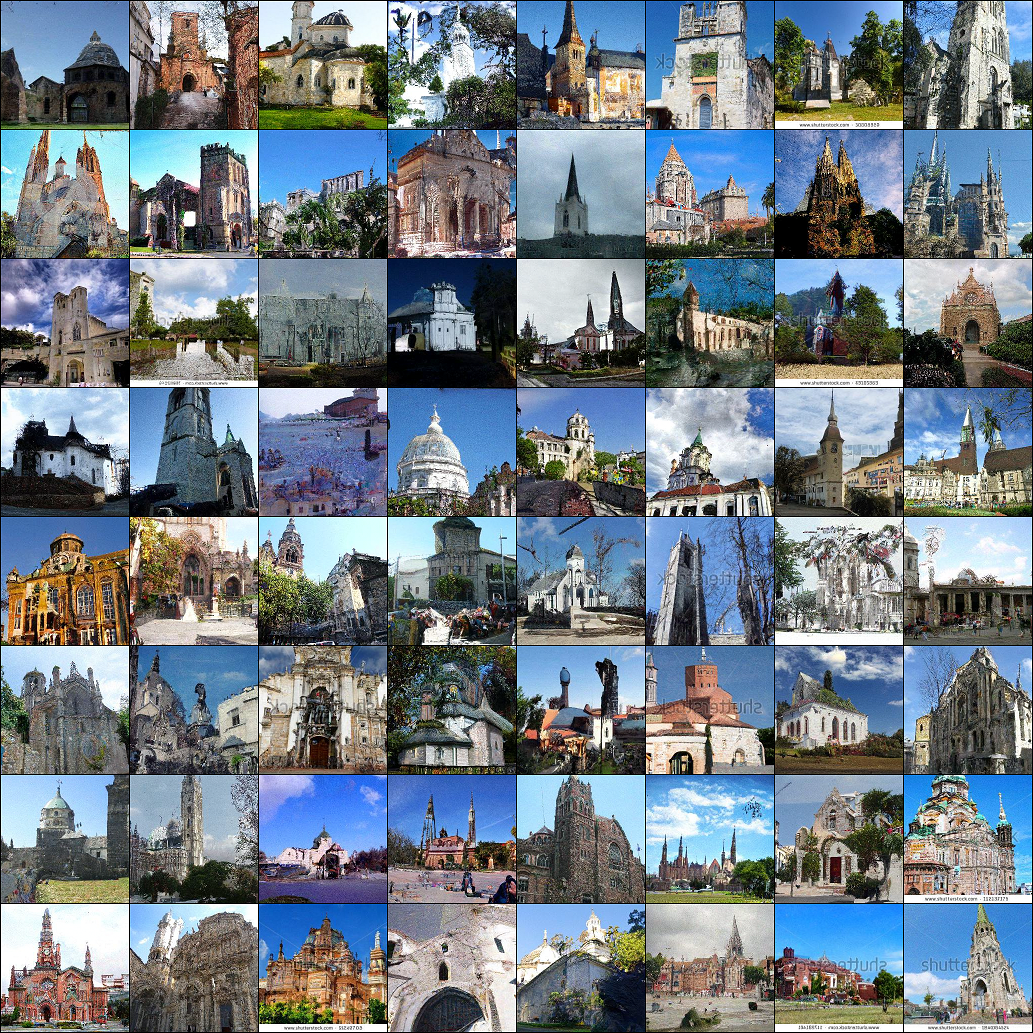}
    \caption{Dynamic-step Extrapolation ($\epsilon=0.10$)} 
  \end{subfigure}
  \caption{VE - LSUN-Church (256x256)}
  \label{fig15} 
\end{figure}

\begin{figure}[ht] 
  \begin{subfigure}[b]{0.5\linewidth}
    \centering
    \includegraphics[width=1\linewidth]{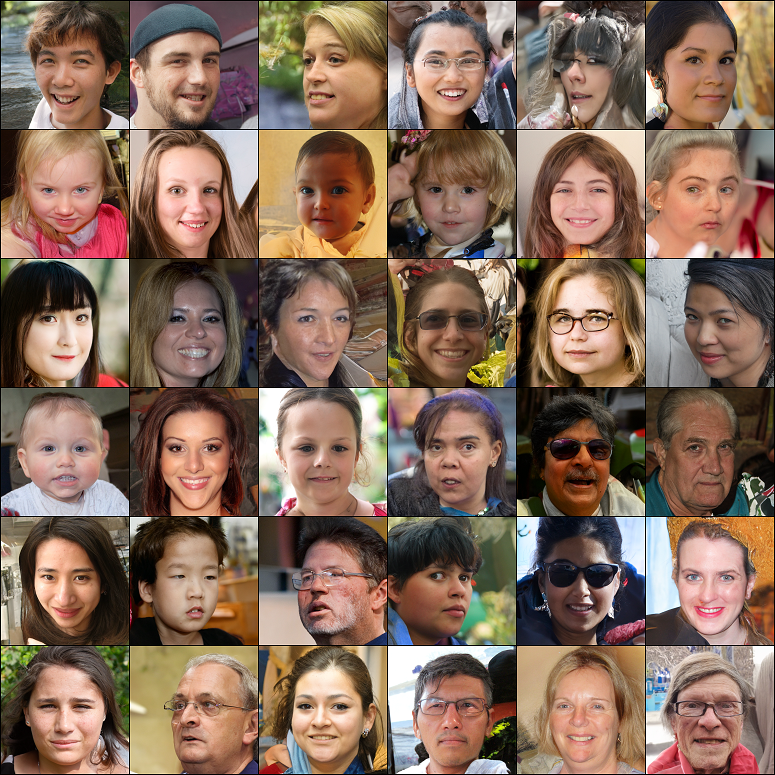}
    \caption{Dynamic-step Extrapolation ($\epsilon=0.01$)} 
  \end{subfigure}
  \begin{subfigure}[b]{0.5\linewidth}
    \centering
    \includegraphics[width=1\linewidth]{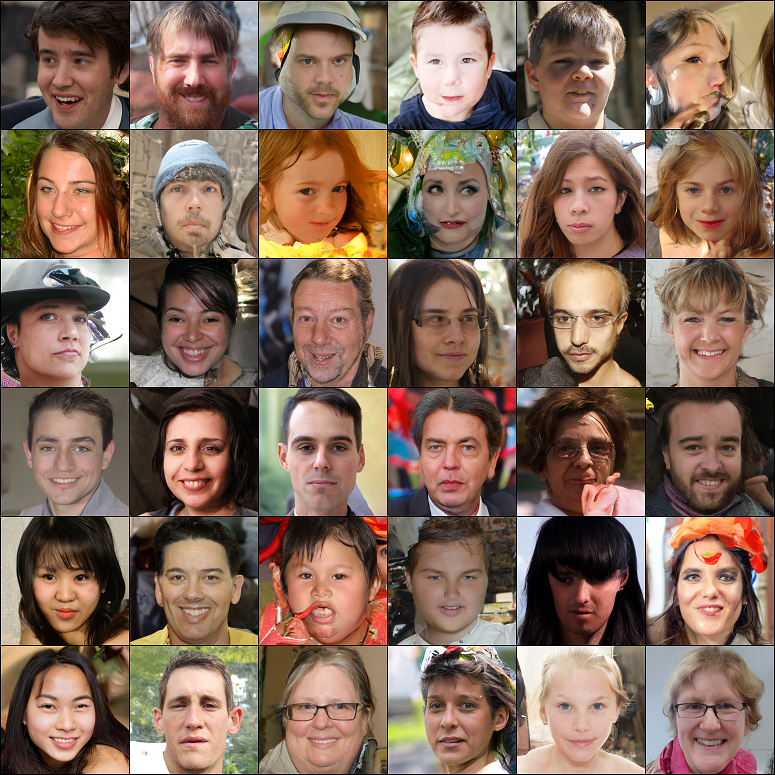}
    \caption{Dynamic-step Extrapolation ($\epsilon=0.02$)} 
  \end{subfigure}
  \begin{subfigure}[b]{0.5\linewidth}
    \centering
    \includegraphics[width=1\linewidth]{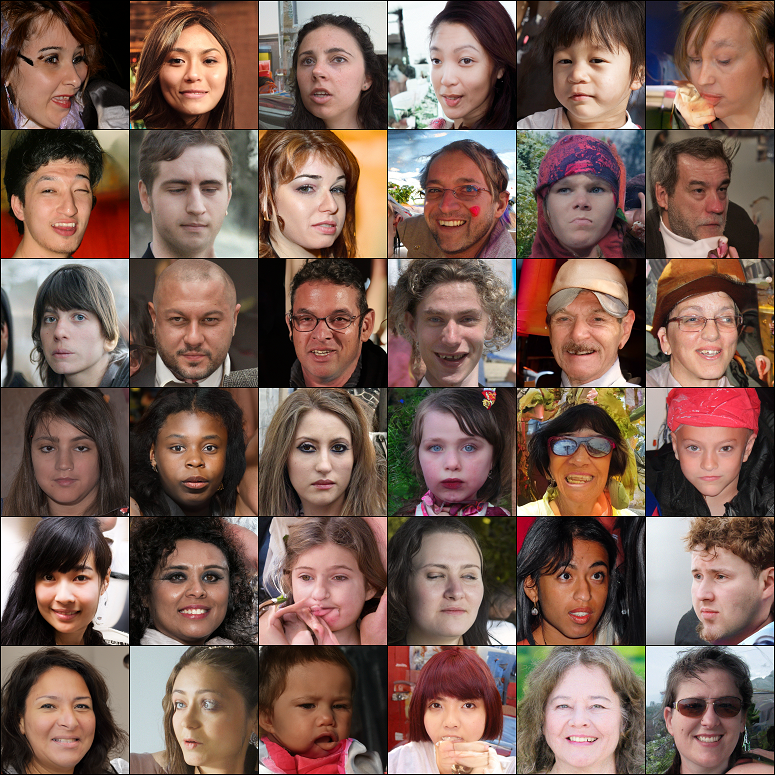}
    \caption{Dynamic-step Extrapolation ($\epsilon=0.05$)} 
  \end{subfigure}
  \begin{subfigure}[b]{0.5\linewidth}
    \centering
    \includegraphics[width=1\linewidth]{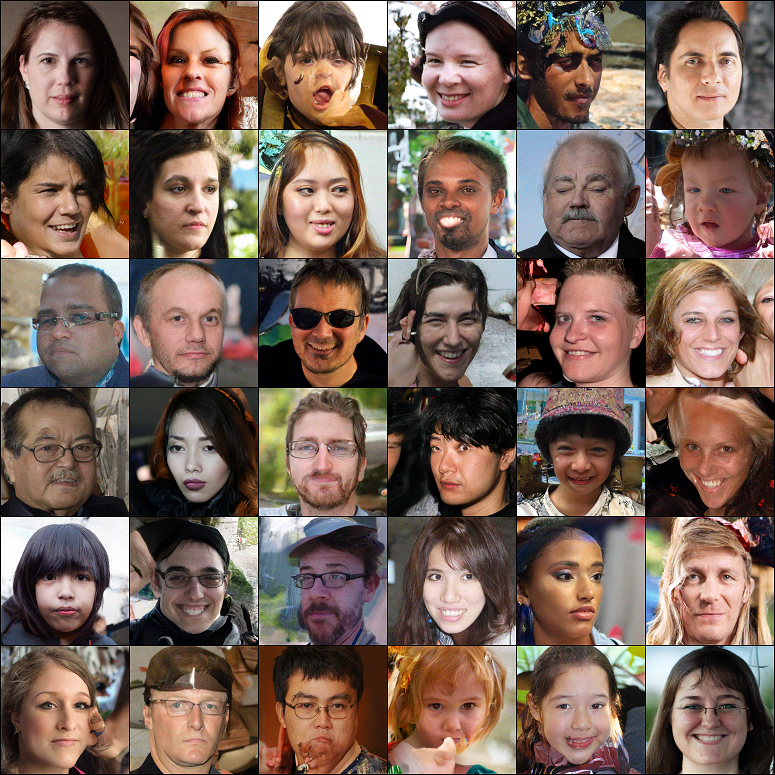}
    \caption{Dynamic-step Extrapolation ($\epsilon=0.10$)} 
  \end{subfigure}
  \caption{VE - FFHQ (256x256)}
  \label{fig16} 
\end{figure}

\end{document}